\documentclass[runningheads]{llncs}

\usepackage[T1]{fontenc}

\usepackage{graphicx}

\usepackage{times}
\usepackage{epsfig}
\usepackage{amsmath}
\usepackage{amssymb}

\usepackage{multirow}
\usepackage{multicol}
\usepackage{caption}
\usepackage{subcaption}
\usepackage{balance}

\usepackage{appendix}
\usepackage{url}
\usepackage{booktabs}

\graphicspath{ {./fig/} }

\newcommand{\B}{\bfseries}

\begin{document}

\title{Transformer-based Cross-Modal Recipe Embeddings with Large Batch Training}
\author{Jing Yang\inst{1} \and
Junwen Chen\inst{1}  \and
Keiji Yanai\inst{1}}
\authorrunning{J. Yang, J. Chen and K. Yanai}
\institute{The University of Electro-Communications, Tokyo, Japan
\email{\{yang-j,chen-j,yanai\}@mm.inf.uec.ac.jp}\\
}
\maketitle              
\begin{abstract}

Cross-modal recipe retrieval aims to exploit the relationships and accomplish mutual retrieval between recipe images and texts, which is clear for human but arduous to formulate. 
Although many previous works endeavored to solve this problem, most works did not efficiently exploit the cross-modal information among recipe data.
In this paper, we present a frustratingly straightforward cross-modal recipe retrieval framework, Transformer-based Network for Large Batch Training~(TNLBT) achieving high performance on both recipe retrieval and image generation tasks, which is designed to efficiently exploit the rich cross-modal information.
In our proposed framework, Transformer-based encoders are applied for both image and text encoding for cross-modal embedding learning. 

We also adopt several loss functions like self-supervised learning loss on recipe text to encourage the model to further promote the cross-modal embedding learning.
Since contrastive learning could benefit from a larger batch size according to the recent literature on self-supervised learning,
we adopt a large batch size during training and have validated its effectiveness.
The experimental results showed that TNLBT significantly outperformed the current state-of-the-art frameworks in both cross-modal recipe retrieval and image generation tasks on the benchmark Recipe1M by a huge margin. We also found that CLIP-ViT performs better than ViT-B as the image encoder backbone.
This is the first work which confirmed the effectiveness of large batch training on cross-modal recipe embedding learning.

\end{abstract}
\section{INTRODUCTION}
Cross-modal recipe retrieval investigates the relation between recipe texts and recipe images to enable mutual retrieval between them. 
Recipe1M dataset~\cite{im2recipe,recipe1m+} is frequently used as a benchmark to evaluate the performance of cross-modal recipe retrieval frameworks. One challenge of the recipe retrieval task is that food images usually contain many non-food parts like plates and different backgrounds as noise. The recipe texts are perplexing, making them difficult to encode, since there are three components in the recipe text: title, ingredients, and instructions,  the last two of which are long and structured texts.
Most existing works focus just on recipe or image embedding learning, though we believe that a framework that focuses on both recipe and image embedding learning is necessary.

As one of the early works Salvador~{\it et al.}~\cite{im2recipe}
proposed Joint Embedding (JE), projecting embeddings from both modalities into a common shared space and minimizing the cosine similarity between them to enable cross-modal recipe retrieval. 
GAN~\cite{gan} was introduced in some recent works~\cite{r2gan,ACME,RDEGAN} for food image synthesis from embeddings to obtain more reliable cross-modal embeddings. By confirming if the same or similar embeddings can be extracted from the food images generated from cross-modal embeddings, the retrieval performance were improved.
Furthermore, modality adversarial loss was introduced in ACME~\cite{ACME} which can narrow the modality gap between image and text embedding.
Inspired by the success of GAN-based image synthesis in recipe retrieval, we also adopt GAN-based image synthesis in this work.
However, regarding recipe text embedding, LSTM~\cite{bi_dire_lstm} was simply applied for encoding recipe texts in these works~\cite{r2gan,ACME,RDEGAN}, 
which sometimes limits leveraging the information in long and structured recipe texts. In order to address this issue, some other methods~\cite{MCEN,icmr20,h-t} focused on improving recipe embedding learning. Authors of MCEN~\cite{MCEN} applied attention mechanism, and Zan {\it et al.}~\cite{icmr20} applied BERT~\cite{BERT} for recipe encoding. 
With the trend of Transformer~\cite{atten} in natural language processing, X-MRS~\cite{x-mrs} and H-T~\cite{h-t} were recently proposed to adopt this technique in recipe retrieval.
Authors of H-T in particular introduced a simple but effective framework with a Transformer-based structure recipe encoder and self-supervised learning, allowing the model to explore complementary information among recipe texts. Inspired by H-T, we adopt a Transformer-based recipe encoder and self-supervised learning for recipe embedding learning. 

In this paper, we propose a frustratingly straightforward Transformer-based framework that uses all the current state-of-the-art techniques. That is, we use a hierarchical transformer architecture recipe encoder for recipe embedding learning, and an adversarial network to investigate the complementary information between recipe and image embedding. In the experiments, we adopt ViT~\cite{vit} and CLIP-ViT~\cite{CLIP-ViT} as the image encoder backbone to validate the effectiveness the proposed framework.
Since contrastive learning benefits from larger batch sizes according to the recent work on self-supervised learning~\cite{simclr} and self-supervised learning and triplet losses are used in the proposed framework, we adopt large batch training in our experiments. 
Furthermore, we conducted extensive experiments and ablation studies to further validate the effectiveness of our proposed framework. We discovered that large batch training was surprisingly effective for Transformer-based cross-modal recipe embedding.
The results showed that our proposed framework outperformed the state-of-the-arts both in recipe retrieval (medR 1.0, R1 56.5 in 10k test set size) and image generation tasks (FID score 16.5) by a large margin.
More specifically, we summarize the contributions of this work as follows:
\begin{enumerate}
 \item We proposed a frustratingly straightforward Transformer-based cross-modal recipe retrieval framework, Transformer-based Network for Large Batch Training (TNLBT), which achieves the state-of-the-art performance on both recipe retrieval and image generation tasks.
 \item We conducted experiments with large batch inspired by the effectiveness of large batch training on contrastive learning~\cite{simclr}. 
 \item Through the comprehensive experiments, we confirmed that the proposed framework outperforms the current state-of-the-arts with a large margin, especially in the case of a large batch size, 768. 
 This is the first work which confirmed the effectiveness of large batch training on cross-modal recipe retrieval as far as we know.
\end{enumerate}

\section{RELATED WORK}

\subsection{Cross-Modal Recipe Retrieval}
Cross-modal retrieval aims to enable mutual retrieval between two different modalities, often image and text. 
The common idea of cross-modal retrieval is to embed features from two different modalities into a shared common space while keeping the distribution of corresponding embedded features close to enable mutual retrieval. A substantial issue in this task is how to narrow the gap between the various modalities~\cite{survey_multimodal}.  

Salvador~{\it et al.}, who was the first to propose the cross-modal recipe retrieval task and the Recipe1M dataset, proposed joint embedding~\cite{im2recipe} to enable cross-modal recipe retrieval. 
This method was modified in AdaMine~\cite{adamine} by using 
a triplet loss~\cite{triplet} to improve the retrieval accuracy. In order to further exploit the information in the Recipe1M dataset, some state-of-the-art techniques like Transformer~\cite{atten} and Generative Adversarial Networks~(GAN)~\cite{gan} are used in recipe retrieval. 
GAN, enables image generation conditioned on recipe embedding in recipe retrieval framework. GAN was introduced in several previous works~\cite{r2gan,ACME,RDEGAN}, which improves retrieval accuracy while enabling image generation conditioned on recipe text. The issues of these previous works are, the retrieval accuracy and quality of generated images are very limited since they only adopted simple LSTM~\cite{bi_dire_lstm} for text embedding learning. However, in our proposed method we obtain generated images of better quality using Transformer.

Rather than incorporating complex networks like GAN, some research~\cite{MCEN,icmr20,h-t} focus on recipe embedding learning to enhance the performance of retrieval tasks. Authors of MCEN\cite{MCEN} introduced cross-modal attention and consistency and Zan {\it et al.}~\cite{icmr20} introduced BERT~\cite{BERT} as a recipe encoder to enable cross-modal retrieval. Authors of X-MRS~\cite{x-mrs} introduced a Transformer~\cite{atten} encoder to gain recipe embedding, further proposed the use of imperfect multilingual translations, and achieved state-of-the-art performances on retrieval tasks.
Salvador {\it et al.}~ proposed a simply but effective framework H-T~\cite{h-t}, to  facilitate the power of Transformer. 
However, authors of these works~\cite{MCEN,icmr20,h-t} just simply applied
bi-directional triplet loss on image and recipe features without image synthesis and reconstruction, which somewhat limited the cross-modal information learning between recipe and image features. 
In order to address this issue, we introduce GAN-based architecture to enhance the cross-modal embedding learning in our architecture in addition to adopting Transformer-based recipe encoders and self-supervised learning on recipe-only samples in the Recipe1M.  

\begin{figure}[t]
\centering
\includegraphics[width =\columnwidth]{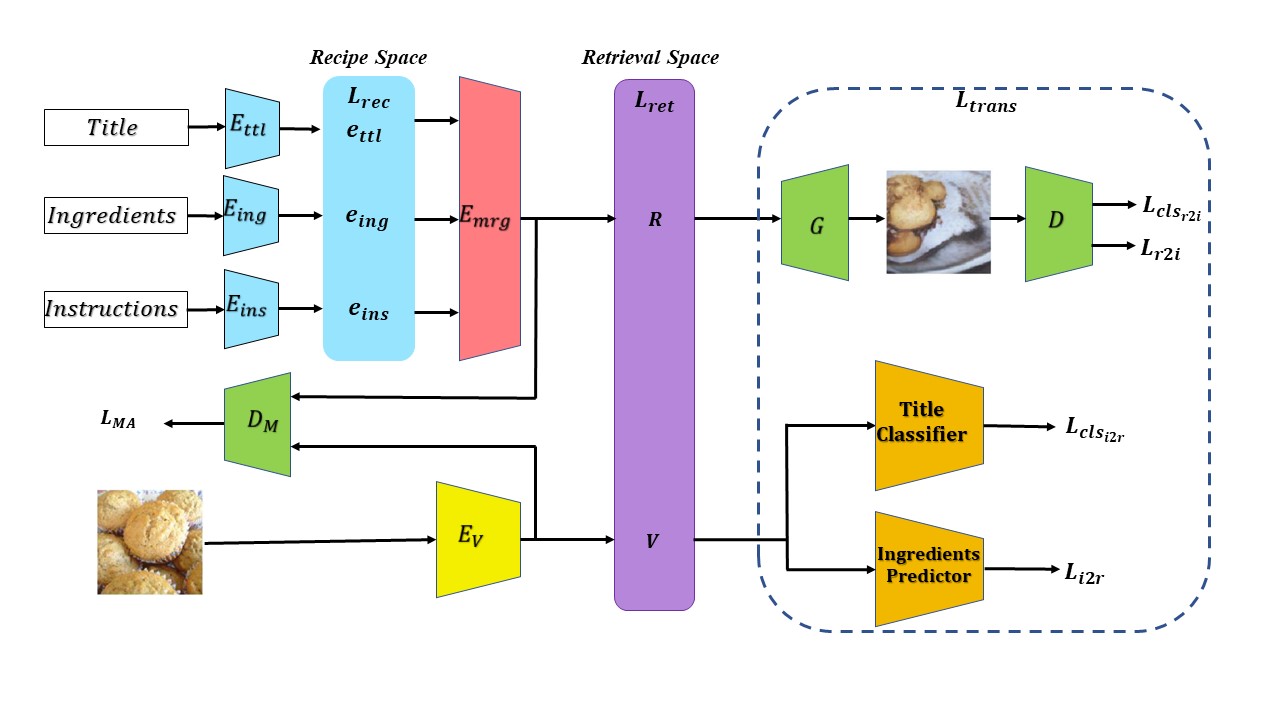}
\caption{\textbf{The architecture of proposed network TNLBT.} Training process is controlled by the four loss functions~$L_{rec},~L_{ret},~L_{MA},~L_{trans}$~. }
\label{network}
\end{figure}

\subsection{Food Image Synthesis}

GAN~\cite{gan} has been introduced and proven its effectiveness for improving recipe retrieval performance in some recent works~\cite{r2gan,ACME,RDEGAN,x-mrs}. 
Authors of R2GAN~\cite{r2gan}, ACME~\cite{ACME} and X-MRS~\cite{x-mrs} applied text-conditioned image synthesis which generates recipe images from recipe embeddings, while authors of RDE-GAN~\cite{RDEGAN} proposed to disentangle image features into dish shape features, which contained only non-recipe information,  and recipe embeddings and to integrate both of them to generate recipe images. Similar to ACME and RDE-GAN, we also leverage GAN in our proposed framework to improve retrieval accuracy while enabling image generation conditioned on recipe text. The biggest difference between our work and the previous works is, we applied Transformer-based encoders to further improve the quality of generated images compared to the previous works~\cite{ACME,RDEGAN}.

\section{Method}

\subsection{Overview}
We propose a frustratingly straightforward Transformer-based framework TNLBT for cross-modal recipe retrieval
which has the hierarchical Transformers for text encoding,
and Vision Transformer (ViT)~\cite{vit} as an image encoder.
Furthermore, we propose to adopt large batch training which is confirmed to be 
beneficial for contrastive training and Transformer-based networks.
Figure \ref{network} shows the architecture of the proposed framework.
Similar to H-T~\cite{h-t}, our proposed framework applies hierarchical Transformer encoders and a triplet loss $L_{rec}$ with self-supervised learning to explore complementary information in the recipe text. 
ViT-B~\cite{vit} or CLIP-ViT~\cite{CLIP-ViT} is used as the backbone of image encoder. The biggest difference between our work and H-T~\cite{h-t} is, we introduce a more sophisticated adversarial network to leverage the learned Transformer-based embeddings. 
Rather than just a bi-directional triplet loss is adopted in H-T for the distance learning of image and recipe embeddings, we further introduce several loss functions to enhance the cross-modal embedding learning and enable image generation conditioned on recipe text. 
Inspired by the success of triplet loss~\cite{triplet}, a triplet loss $L_{ret}$ for distance learning in image-recipe retrieval is applied in the retrieval space. 
In order to mitigate the modality gap problem, we use the modality alignment loss function $L_{MA}$ as in ACME~\cite{ACME}. Furthermore, we introduce image-to-recipe and recipe-to-image information recover, and a translation consistency loss function $L_{trans}$ to keep modality-specific information. We expect that the embeddings learned in the above process correctly retains the original information and enables image generation from recipe embeddings. 
Finally the overall loss function can be formulated as follows:
\begin{equation} \label{final_loss}
L_{total} = \lambda_1 L_{rec} +\lambda_2 L_{MA} + \lambda_3 L_{trans} + L_{ret} ,
\end{equation}
where $\lambda_1,~\lambda_2,~\lambda_3$ are hyperparameters to control the loss balance.

\subsection{Recipe Encoder and Self-Supervised Learning}
The purpose of the recipe encoder, $E_R$~(even though we adopt several encoders here, we note these as a whole recipe encoder $E_R$), is to encode recipe text, $\textbf{t}$, to recipe embeddings, $\textbf{R}$.
In order to better explore and make use of the huge information among recipe text with components of title, ingredients, and instructions, we adopt the hierarchical Transformers as the recipe encoder similar to H-T~\cite{h-t}. Note that we introduced an adversarial network for image synthesis and modality alignment better to exploit the correlation in image-recipe pairs, while H-T simply applied a bi-directional triplet loss function to image-recipe pairs.

For the three components~(title, ingredients, instructions) in the recipe texts, the proposed framework TNLBT encodes them separately using hierarchical Transformer encoders~($E_{ttl}, E_{ing}, E_{ins}$), with 2 layers and 4 attention heads, to obtain embeddings of title, ingredients and instructions ($e_{ttl}, e_{ing}, e_{ins}$) initially.
Next, we apply self-supervised recipe loss $L_{rec}$ as in H-T, on $e_{ttl}, e_{ing}, e_{ins}$ to explore the complementary information among them. The introduction of this loss allows us to use recipe-only image-recipe pairs to further leverage the complementary information among the three components of the recipe text. Finally, we apply a merging encoder $E_{mrg}$ to project three components embeddings ($e_{ttl}, e_{ing}, e_{ins}$) to an unified recipe embedding, $\textbf{R}$.

\subsection{Image Encoder}
The purpose of the image encoder $E_V$ is to encode recipe image \textbf{i} to image embedding $\textbf{V}$. We adopt the 
base size model of Vision Transformer (ViT-B)~\cite{vit} pre-trained on ImageNet-21k~\cite{imageNet} as the image encoder backbone. In order to better leverage the benefit of large batch training, we also adopt CLIP-ViT~\cite{CLIP-ViT} as image encoder backbone where the proposed framework achieved superior performance on retrieval task. 

\subsection{Modality Alignment Loss} 
For the purpose of narrowing the gap between recipe and image embeddings, which often results in bad generalization or slow convergence, we applied modality alignment loss $L_{MA}$ as in ACME~\cite{ACME}.
We adopt a discriminator $D_M$ here to achieve this goal, which aligns the distribution of recipe embeddings $R = E_R(\textbf{t})$ and image embeddings $V = E_V(\textbf{i})$ during training.
An adversarial loss is applied by controlling the recipe and image embeddings aligned to each other resulting in the discriminator $D_M$ cannot distinguish the source of the given embeddings empirically. Following ACME, we also adopt WGAN-GP~\cite{wgangp} here and the loss function is formulated as follows:
\begin{multline} \label{loss_MA}
L_{MA} = \mathbb{E}_{\textbf{i} \sim p(i)} [\log(1 - D_M(E_V(\textbf{i})))] +
\mathbb{E}_{\textbf{t} \sim p(t)} [\log(1 - D_M(E_R(\textbf{t})))]
\end{multline}

\subsection{Retrieval Loss} 
\label{retrieval_learning_section}
Following the success of triplet loss~\cite{triplet} in the recent works~\cite{adamine,ACME,RDEGAN,h-t}, we adopt a triplet loss for distance learning. Here we obtain the recipe and image embeddings $ R=E_R(\textbf{t}), V=E_V(\textbf{i})$.  We obtain an anchor recipe embedding $R_a$ and an anchor image embedding $V_a$, and obtain a negative sample with subscript $n$ and a positive sample with subscript $p$ to process this distance learning. The loss function is formulated as follows:
\begin{multline} \label{loss_tri}
L_{ret} = \sum_V{[d(V_a, R_p) - d(V_a, R_n) + \alpha}]_+ +\\
    \sum_R{[d(R_a, V_p) - d(R_a, V_n) + \alpha}]_+~,
\end{multline}
where margin $\alpha=0.3$, $d(.) $ is the Euclidean distance, $[x]_+=max(x, 0)$.
In addition, we adopt a hard sample mining strategy~\cite{hard_sample_mining} to further facilitate the distance learning.

\subsection{Translation Consistency Loss} 
The learned embeddings are useful for training but sometimes lost the modality-specific information which is meaningful and important. To alleviate this information loss, we adopt translation consistency loss following ACME~\cite{ACME}, which ensures the learned embeddings preserve the original information across modalities. We accomplish this goal by forcing the recipe and image embeddings to recover the information in the other modalities: recipe image generation and ingredients prediction conditioned on recipe embedding and image embedding respectively. 
Hence, the total translation consistency loss is composed of the losses for the recipe and the image as follows:
\begin{equation} \label{loss_trans_total}
L_{trans} = L_{trans_r} + L_{trans_i}
\end{equation}

\subsubsection{\B Image Generation from Recipe Embeddings}
In order to preserve the modality-specific information in recipe embeddings, we aim to recover the information in image modality to ensure this property, and we set two losses as for two-fold goal: (1) we expect the generated image is as realistic as possible, (2) we expect the generated realistic image matches the target recipe. 
To accomplish these two goals, we adopt GAN~\cite{gan} to generate images from recipe embeddings and introduce a loss $L_{r2i}$ for goal (1) and a loss $L_{cls_{r2i}}$ for goal (2) following ACME~\cite{ACME}.
During training, the discriminator $D_{r2i}$ is used to distinguish the generated image and real image, and the generator $G$ is used to generate food images which is conditioned on the recipe embedding $FC(E_R(r))$. The loss $L_{r2i}$ for ensuring the generated images realistic is formulated as follows:
\begin{multline}\label{loss_g_r2i}
L_{r2i} = \mathbb{E}_{\textbf{i} \sim p(i)} [\log(1 - D_{r2i}(\textbf{i}))] +
\mathbb{E}_{\textbf{t} \sim p(t)} [\log(1 - D_{r2i}(G(\textbf{FC}(E_R(\textbf{t})))))]
\end{multline}
While ensuring the generated image is as realistic as possible, we introduce loss $L_{cls_{r2i}}$ to ensure that the generated image matches the target recipe. During training, a classifier $cls_{r2i}$ is used to encourage the generator to generate a food image with the corresponding food category to recipe embedding. $L_{cls_{r2i}}$ is simply a cross-entropy loss. Combining these two loss functions for two goals separately, the translation consistency loss for the recipe is formulated as follows:
\begin{equation} \label{loss_trans_r2i}
L_{trans_r} = L_{r2i} + L_{cls_{r2i}}
\end{equation}

\subsubsection{\B Classification and Prediction from Image Embeddings}
We adopt ingredients prediction and title classification on image embeddings here for ensuring the translation consistency of images. 
We leverage a multi-label network here to predict the ingredients from image embeddings using a 4,102-d one-hot vector representing the existence of 4,102 different ingredients. We denote this multi-label objective as $L_{i2r}$.

The same as ensuring the generated image is with the correct food category in defining $L_{cls_{r2i}}$, we propose to ensure the predicted ingredients are from the food with the correct category, which means the image embeddings can be classified into the correct food category. We use $L_{cls_{i2r}}$ here to make sure the image embeddings can be classified into one correct food category of 1,047 recipe categories, where $L_{cls_{i2r}}$ is a cross-entropy loss. 
Combining these two loss functions, the translation consistency loss for the image is formulated as follows:
\begin{equation} \label{loss_trans_i2r}
L_{trans_i} = L_{i2r} + L_{cls_{i2r}}
\end{equation}

\section{EXPERIMENTS}
In this section, we present the experiments to validate the effectiveness of our proposed framework TNLBT both in recipe retrieval and image generation tasks, including ablation studies and  comparison with the previous works. Extensive experiments are also performed to further validate the effectiveness of TNLBT.

\begin{table*}[t]
  \caption{Comparison with the existing works on retrieval performance. The performance of recipe retrieval is evaluated on the criteria of medR$(\downarrow)$ and R@\{1,5,10\}$(\uparrow)$. Especially all the R@K metrics were all improved by around 10.0\% in TNLBT-C compared to TNLBT-V.}
  \label{retrievalRsl}
\resizebox{\textwidth}{!}{%
  \begin{tabular}{l|cccc|cccc|cccc|cccc}
    \toprule
    \multirow{3}{*}{} &
      \multicolumn{8}{c}{1k} &
      \multicolumn{8}{c}{10k} \\
      \cline{2-17}
      &
      \multicolumn{4}{c}{Image-to-Recipe} &
      \multicolumn{4}{c}{Recipe-to-Image} &
      \multicolumn{4}{c}{Image-to-Recipe} &
      \multicolumn{4}{c}{Recipe-to-Image} \\
      \cline{2-17}
      & {medR} & {R@1} & {R@5} & {R@10} & {medR} & {R@1} & {R@5} & {R@10}
      & {medR} & {R@1} & {R@5} & {R@10} & {medR} & {R@1} & {R@5} & {R@10}\\
      \midrule
        JE~\cite{im2recipe} & 5.2& 24.0& 51.0& 65.0& 5.1& 25.0& 52.0& 65.0& 41.9& {-} & {-}& {-} & 39.2& {-}& {-}& {-}\\
        R2GAN~\cite{r2gan} & 2& 39.1& 71.0& 81.7& 2& 40.6& 72.6& 83.3& 13.9& 13.5& 33.5& 44.9& 12.6& 14.2& 35.0& 46.8  \\
        MCEN~\cite{MCEN} &  2& 48.2& 75.8& 83.6& 1.9& 48.4& 76.1& 83.7& 7.2& 20.3& 43.3& 54.4& 6.6& 21.4& 44.3& 55.2 \\
        ACME~\cite{ACME} & 1& 51.8& 80.2& 87.5& 1& 52.8& 80.2& 87.6 & 6.7& 22.9& 46.8& 57.9& 6& 24.4& 47.9& 59.0\\
        SCAN~\cite{scan} & 1& 54.0& 81.7& 88.8& 1& 54.9& 81.9& 89.0& 5.9& 23.7& 49.3& 60.6& 5.1& 25.3& 50.6& 61.6 \\
        IMHF~\cite{IMHF} &  1& 53.2& 80.7& 87.6& 1& 54.1& 82.4& 88.2& 6.2& 23.4& 48.2& 58.4& 5.8& 24.9& 48.3& 59.4 \\
        RDEGAN~\cite{RDEGAN} &  1& 59.4& 81.0& 87.4& 1& 61.2& 81.0& 87.2 & 3.5& 36.0& 56.1& 64.4& 3& 38.2& 57.7& 65.8 \\
        H-T~\cite{h-t} & 1& 60.0& 87.6& 92.9& 1& 60.3& 87.6& 93.2& 4& 27.9& 56.4& 68.1& 4& 28.3& 56.5& 68.1   \\
        X-MRS~\cite{x-mrs} & 1& 64.0& 88.3& 92.6& 1& 63.9& 87.6& 92.6& 3& 32.9& 60.6& 71.2& 3& 33.0& 60.4& 70.7 \\
    \bottomrule
    TNLBT-V &\underline{1}&\underline{75.1}&\underline{92.3}&\underline{95.3}&\underline{1}&\underline{75.2}&\underline{92.5}&\underline{95.4} & 
    \underline{2}&\underline{48.0}&\underline{73.7}&\underline{81.5}&\underline{2}&\underline{48.5}&\underline{73.7}&\underline{81.5}\\
    TNLBT-C &\B1&\B81.0&\B95.2&\B97.4& \B1&\B80.3&\B95.2&\B97.4&\B1&\B56.5&\B80.7&\B87.1 &\B1&\B55.9&\B80.1&\B86.8\\
    
    \bottomrule
  \end{tabular}}

\end{table*}

\begin{table*}[!t]
\begin{minipage}{0.5\linewidth}
\centering
\caption{Comparison with the existing works on image generation. The performance of image generation is evaluated on the criteria of FID$(\downarrow)$. }
\label{fid_score}

\resizebox{0.65\linewidth}{!}{%
\setlength{\tabcolsep}{20pt}
\begin{tabular}{lc}
    \toprule
      Method & FID\\
      \hline
      ACME~\cite{ACME} & 30.7 \\
      CHEF~\cite{CHEF} & 23.0 \\
      X-MRS~\cite{x-mrs} & 28.6 \\
      \bottomrule
      TNLBT-V & \underline{17.9} \\
      TNLBT-C & \textbf{16.5} \\

    \bottomrule
    
  \end{tabular}
}

  \caption{Evaluation of the importance of different components on image-to-recipe retrieval in TNLBT. * means training without recipe-only data.}
  \label{loss_comb_comparison_rsl}
\resizebox{\linewidth}{!}{%
  \begin{tabular}{l|cccc}
    \toprule
      Applied Components & {medR} & {R@1} & {R@5} & {R@10} \\
    \midrule
    $L_{ret}$&2.0&43.4&70.3&79.4\\
    $L_{ret}$+$L_{MA}$&2.0&41.1&68.0&77.3\\
    $L_{ret}$+$L_{MA}$+$L_{rec}$&2.0&43.2&70.0&79.0\\
    $L_{ret}$+$L_{MA}$+$L_{rec}$+$L_{trans_i}$&2.0&43.5&70.3&79.0\\
    $L_{ret}$+$L_{MA}$+$L_{rec}$+$L_{trans_r}$&2.0&44.5&71.3&80.2\\
    $L_{ret}$+$L_{rec}$+$L_{trans_r}$+$L_{trans_i}$&2.0&43.5&70.7&79.5\\
    $L_{ret}$+$L_{MA}$+$L_{rec}$+$L_{trans_r}$+$L_{trans_i}$* &2.0&43.1&69.8&78.7\\
    $L_{ret}$+$L_{MA}$+$L_{rec}$+$L_{trans_r}$+$L_{trans_i}$ & 2.0& 44.3& 70.9& 79.7\\
    TNLBT-V~(large batch training) &2.0&48.0&73.7&81.5\\
    TNLBT-C~(CLIP-ViT applied) & \B1.0&\B56.5&\B80.7&\B87.1 \\
    \bottomrule
    
  \end{tabular}
  }

\end{minipage}
\hfill
\begin{minipage}{0.48\linewidth}
\centering
\caption{Comparison on image-to-recipe retrieval performance with different batch sizes adopted in TNLBT. Results are reported on rankings of size 10k.}
\label{batchsize__vit}
\resizebox{\linewidth}{!}{%
  \begin{tabular}{l|cccc|cccc}
    \toprule
    \multirow{2}{*}{}
       &
      \multicolumn{4}{c}{\B TNLBT-V} &
      \multicolumn{4}{c}{\B TNLBT-C} \\
      \hline
      \#batch & {medR} & {R@1} & {R@5} & {R@10}  &{medR} & {R@1} & {R@5} & {R@10}\\
    \midrule
    64 &3.0&36.6&64.3&74.3 &2.0&48.0&75.4&83.9\\
    128 &2.0&40.9&68.3&77.6&1.4&50.1&77.1&84.9\\
    256 & 2.0& 44.3& 70.9& 79.7&1.0&53.5&79.1&86.3\\
    512 & 2.0& 47.1& 73.4& \B81.6&1.0&55.9&80.4&86.9\\
    768&\B2.0&\B48.0&\B73.7&81.5&\B1.0&\B56.5&\B80.7&\B87.1\\
    1024&2.0&47.5&73.3&81.2&1.0&56.0&79.8&86.5\\
    \bottomrule
  \end{tabular}}


\vspace{5.5mm}
\caption{Evaluation on image generation in different batch sizes on the criteria of FID$(\downarrow)$.}
\label{fid_score_batch}
\resizebox{\linewidth}{!}{%
    \setlength{\tabcolsep}{10pt}
      \begin{tabular}{l|cc}
        \toprule
          \#batch &FID~(TNLBT-V) &FID~(TNLBT-C)\\
          \hline
          64&  \B 17.9& \B 16.5\\
          128&   19.3& 21.9\\
          256&  27.1& 30.1\\
          512& 34.4& 48.6\\
          768 &  46.4& 69.7\\
        \bottomrule
      \end{tabular}
}

\end{minipage}
\end{table*}


\subsection{Implementation Details}

\noindent\textbf{Data set.}\quad We evaluated the performance of our proposed method on Recipe1M~\cite{im2recipe,recipe1m+} following the previous works. Following the official dataset splits, we used 238,999 image-recipe pairs for training, 51,119 pairs for validation, and 51,303 pairs for testing. 

\noindent\textbf{Evaluation Metrics.}\quad Following previous works, we evaluated retrieval performance with the median rank (medR) and R@\{1,5,10\} on two different test set sizes 1k and 10k. We reported the average metrics of 10 groups which are randomly chosen from the test set.
We also evaluated the performance of image generation conditioned on recipe embeddings using Fréchet Inception Distance (FID)~\cite{fid_paper} score following previous works~\cite{RDEGAN,x-mrs} measuring the similarity of the distribution between generated images and ground-truth images.

\noindent\textbf{Training Details.}\quad The objective is to minimize the total loss function Eq. \ref{final_loss}. We empirically decided the hyperparameters in Eq. \ref{final_loss} as follows: $\lambda_1=0.05,~\lambda_2=0.005,~\lambda_3=0.002$.
For ViT-B~\cite{vit} image encoder backbone framework, We trained the model for 50 epochs using Adam~\cite{adam} with a learning rate of $10^{-4}$. For CLIP-ViT~\cite{CLIP-ViT} image encoder backbone framework, we first trained the model for 20 epochs while freezing the image encoder with a learning rate of $10^{-4}$ and then we trained the model for another 100 epochs with a learning rate of $10^{-6}$ where the image encoder were not frozen. 
Since contrastive learning benefits from larger batch sizes and more training steps compared to supervised learning~\cite{simclr}, batch size was set to 768 during training.

\subsection{Cross-Modal Recipe Retrieval}
We evaluated and compared the performance of the proposed framework TNLBT with the previous works in Table \ref{retrievalRsl}. TNLBT with ViT-B~\cite{vit} image encoder backbone (TNLBT-V) outperforms all the existing works across all metrics with large margins. We also tested the performance of our proposed framework with CLIP-ViT~\cite{CLIP-ViT} image encoder backbone (TNLBT-C), which outperforms all the existing works across metrics with even much larger margins.
When the image encoder backbone was changed from ViT-B to CLIP-ViT, all the R@K metrics were all improved by around 10.0\%, and medR on 10k test set size was improved to $1.0$, which has never been achieved before.

\begin{figure*}[!t]
\centering
\includegraphics[width=\textwidth]{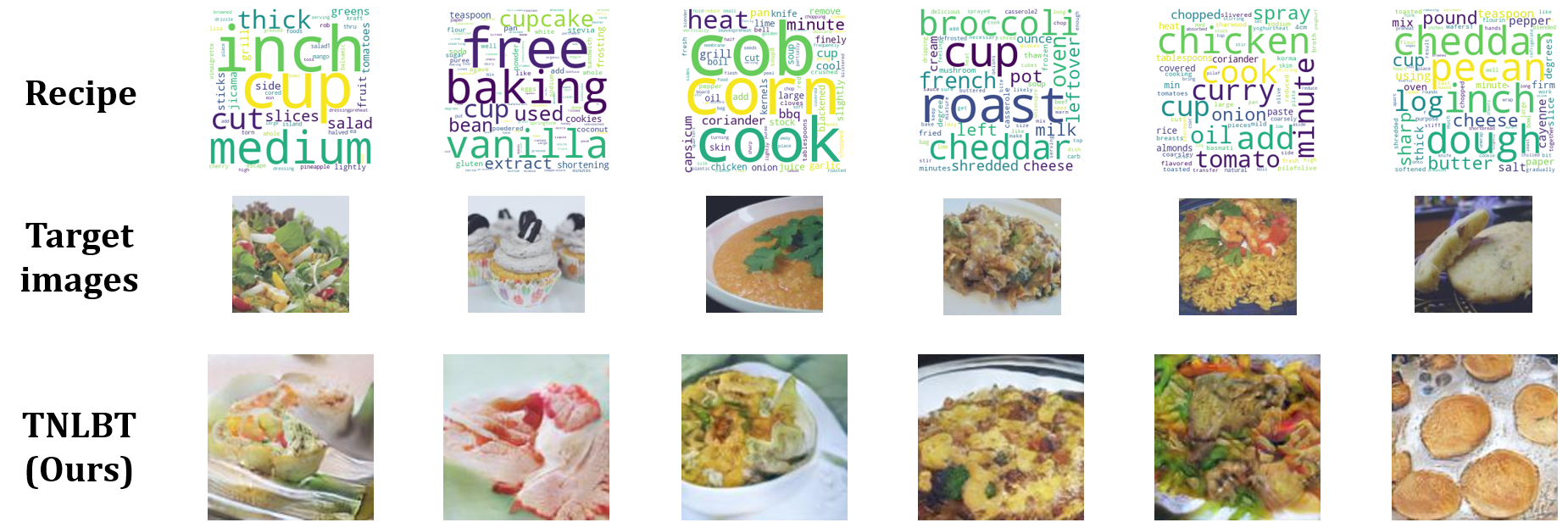}
\caption{Generated images conditioned on different recipe embeddings.
    The first row are recipes used to generate images, the second row are the ground-truth images which recipe embeddings aim to recover. The third row are the generated images.}
\label{recipe_to_image_rsl}
\end{figure*}

\subsection{Image Generation}
The proposed model generates images of $128\times 128$ resolutions conditioned on recipe embeddings. We computed the FID~\cite{fid_paper} score to evaluate the image generation performance of our proposed method\footnote{We used the open source code on \url{https://github.com/mseitzer/pytorch-fid}}.
The results are reported in Table \ref{fid_score}, where TNLBT outperformed the previous works~\cite{ACME,CHEF,x-mrs} with a large margin on image generation\footnote{We borrow the FID scores of CHEF~\cite{CHEF} and ACME~\cite{ACME} reported in X-MRS~\cite{x-mrs}.}. 
Figure \ref{recipe_to_image_rsl} shows qualitative results on image generation conditioned on recipe embeddings, showing that TNLBT could generate appropriate recipe images corresponding to the given ingredients in recipes.

\subsection{Ablation Studies}
We also performed ablation studies to evaluate the importance of the applied components in Table \ref{loss_comb_comparison_rsl}. When $L_{MA}$ was applied, the accuracy decreased incrementally while with more loss functions applied the accuracy increased again. 
However, when only $L_{MA}$ was taken, the retrieval accuracy decreased compared to the combination of all loss functions. Hence we empirically think there is a complementary relationship among $L_{MA}$ and the other loss terms except $L_{ret}$, contributing to the retrieval performance. The loss combination without $L_{trans_i}$ applied outperforms all the other combinations by an incremental margin. Finally, when the proposed large batch training strategy and CLIP-ViT~\cite{CLIP-ViT} applied, the retrieval accuracy was further improved.

\subsection{Batch Size in Cross-Modal Embedding Learning}
Since contrastive learning benefits from large batch size was reported in \cite{simclr}, we performed extensive experiments here to validate the influence of batch size on both recipe retrieval and image generation, where we are the first to validate this influence.

\noindent\textbf{Recipe Retrieval.}\quad Table~\ref{batchsize__vit} shows the image-to-recipe retrieval performance of TNLBT with different batch sizes. 
Retrieval performance was improved substantially when the batch size increased from 64 to 512, indicating the importance of adopting large batch during training. The improvement by increasing batch size is limited afterwards and the performance was hurt incrementally when batch size increased to 1024. 

\noindent\textbf{Image Generation.}\quad We also investigated the influence of batch size on image generation performance in Table \ref{fid_score_batch}, where the performance of image generation hurt with batch size increasing, while the performance of recipe retrieval improved according to Table~\ref{batchsize__vit}. Hence, we emperically believe that batch size serves as a trade-off here.

\section{CONCLUSIONS}
In this research, we proposed a frustratingly straightforward
Transformer-based Network for Large Batch Training~(TNLBT)
using the Hierarchical Transformer-based recipe encoder, the Vision Transformer-based image encoder, and a sophisticated adversarial network for cross-modal recipe embedding learning. We further adopted self-supervised learning to investigate the complementary information in recipe texts.
Through the experiments, it was confirmed that TNLBT outperformed 
the state-of-the-arts on both cross-modal recipe retrieval and food image generation tasks by large margins. We also found that CLIP-ViT~\cite{CLIP-ViT} achieved much better performance than ViT-B~\cite{vit} as an image encoder backbone in TNLBT.
With the extensive experiments, we found that the retrieval performance could benefit from a large batch size while the performance of image generation conditioned on recipe embeddings sometimes got hurt by a large batch size, where we were the first to validate the influence of batch size in recipe retrieval and image generation tasks.

\newpage
{\small
\balance
\bibliographystyle{splncs04}
\bibliography{main}
}
\end{document}